\documentclass[conf]{new-aiaa}
\usepackage[utf8]{inputenc}

\usepackage{graphicx}
\usepackage{amsmath}
\usepackage[version=4]{mhchem}
\usepackage{siunitx}
\usepackage{longtable,tabularx}
\usepackage{xcolor}
\usepackage{subcaption}
\usepackage{multirow}
\usepackage{soul}
\usepackage{gensymb}
\usepackage{algorithm}
\usepackage[noend]{algpseudocode}

\newcommand\blfootnote[1]{%
  \begingroup
  \renewcommand\thefootnote{}\footnote{#1}%
  \addtocounter{footnote}{-1}%
  \endgroup
}

\title{Tradeoffs When Considering Deep Reinforcement Learning for Contingency Management in Advanced Air Mobility }

\author{Luis E. Alvarez\footnote{Luis.Alvarez@ll.mit.edu, Technical Staff, Surveillance Systems Group, 244 Wood Street, Lexington, MA 02421, AIAA Senior Member}, Marc W. Brittain\footnote{Formerly MIT Lincoln Laboratory, Surveillance Systems Group, 244 Wood Street, Lexington, MA 02421, AIAA Senior Member}}
\affil{Massachusetts Institute of Technology Lincoln Laboratory, Lexington, MA, 02421, USA}

\author{Steven D. Young \footnote{Aerospace Research Engineer, Safety Critical Avionics System Branch, MS 234, AIAA Fellow.}}
\affil{NASA Langley Research Center, Hampton, VA, 23681, USA}

\begin{document}
\blfootnote{DISTRIBUTION STATEMENT A. Approved for public release. Distribution is unlimited. This material is based upon work supported by the National Aeronautics and Space Administration under Air Force Contract No. FA8702-15-D-0001. Any opinions, findings, conclusions or recommendations expressed in this material are those of the author(s) and do not necessarily reflect the views of the National Aeronautics and Space Administration. \copyright~2024 Massachusetts Institute of Technology and the United States Government as represented by the Administrator of the National Aeronautics and Space Administration. Delivered to the U.S. Government with Unlimited Rights, as defined in DFARS Part 252.227-7013 or 7014 (Feb 2014). Notwithstanding any copyright notice, U.S. Government rights in this work are defined by DFARS 252.227-7013 or DFARS 252.227-7014 as detailed above. Use of this work other than as specifically authorized by the U.S. Government may violate any copyrights that exist in this work.}
\maketitle

\begin{abstract}
Air transportation is undergoing a rapid evolution globally with the introduction of Advanced Air Mobility (AAM) and with it comes novel challenges and opportunities for transforming aviation. As AAM operations introduce increasing heterogeneity in vehicle capabilities and density, increased levels of automation are likely necessary to achieve operational safety and efficiency goals. This paper focuses on one example where increased automation has been suggested. Autonomous operations will need contingency management systems that can monitor evolving risk across a span of interrelated (or interdependent) hazards and, if necessary, execute appropriate control interventions via supervised or automated decision making. Accommodating this complex environment may require automated functions (autonomy) that apply artificial intelligence (AI) techniques that can adapt and respond to a quickly changing environment. This paper explores the use of Deep Reinforcement Learning (DRL) which has shown promising performance in complex and high-dimensional environments where the objective can be constructed as a sequential decision-making problem. An extension of a prior formulation of the contingency management problem as a Markov Decision Process (MDP) is presented and uses a DRL framework to train agents that mitigate hazards present in the simulation environment. A comparison of these learning-based agents and classical techniques is presented in terms of their performance, verification difficulties, and development process.  
\end{abstract}

\section{Introduction}
\lettrine{T}{he} air transportation system is currently undergoing a rapid evolution with the introduction of novel concepts and the demand for an increase in operational efficiency. One of those novel concepts is Advanced Air Mobility (AAM), which foresees the transport of cargo and passengers across cities, as well as communities currently under-served by aviation. These AAM operations may include electric vertical take-off and landing (eVTOL) aircraft, high-density operations, and a combination of piloted (i.e., remote or on-board) and autonomous flights. Across the world, organizations such as the FAA, NASA, and EASA expect these operations to increase in density from a single operation per hour to over 100 simultaneous operations per hour over a local region~\cite{nasa_conops,faa_conopsv2, barrado2020u, japanAAMConops}. The FAA and NASA refer to the increasing phases of complexity for AAM operations as Urban Air Mobility (UAM) Maturity Levels (UML). Low-density human piloted airspace operations are classified as UML 1, while highly autonomous and dense operations are described as UML 4 and beyond. Evident in NASA's and the FAA's concepts of operations, is an expectation that some types of operations will transition from human piloted to fully autonomous operations. Artificial Intelligence (AI) techniques are viewed as one way to enable higher levels of autonomy because the operations will need to make safety critical decisions within dynamic, high-density environments and in the presence of complex hazards, where other algorithmic approaches are inflexible and difficult to scale. To abate the expected types of hazards, autonomous systems will need to implement Contingency Management (CM), a capability described by NASA within the In-Time Aviation Safety Management Systems (IASMS) concept~\cite{ellis2021concept}. Contingency management is the ability to (1) plan a set contingencies prior to and if necessary during flight, (2) decide which contingency to execute based on available information and context (e.g., aircraft or system states, hazard level), and (3) execute the contingency\footnote{Re-planning may be required at various rates depending on several factors (including the changing flight context)}.

Contingencies may be defined at various levels of detail, for example, (1) a procedure to be followed by a pilot or air traffic controller; (2) a change to the destination (i.e., divert); (3) a revised set of waypoints to traverse; and (4) a change in airspeed, altitude, or heading. The latter three are often wrapped into path planning algorithms; which essentially create updated trajectories in real time. Thus, the prospect of aircraft entering unsafe states can be mitigated by using path planning algorithms or playbooks constructed from Mixed Integer Linear Programming (MILP) Optimization. Within these approaches, a fixed time horizon and dynamic constraints are used to explore optimal paths. Fundamental to prior work is the use of algorithms such as Dijkstra, A*, Probabilistic Roadmap, Potential Fields, Rapidly-Exploring Random Tree (RRT) with approximation of the trajectories through discrete nodes (i.e., grid or graph nodes), Dubins paths, or trajectories generated through a fully defined dynamics model~\cite{dreyfus1969appraisal, lozano1990spatial, barraquand1992numerical, melchior2007particle, lugo2014dubins, 9594498, HAGHIGHI2022108453}. Considering uncertainty within the path planning algorithms poses several challenges. For example, A* and Dijkstra rely on deterministic cost functions and heuristics to provide their convergence guarantees and adaptations for uncertainty can reduce the quality of solutions. Potential fields are rooted on determinism as they rely on scalar fields with sources and sinks representing objects in the environment (e.g., buildings, intruding aircraft, destination, and origin sites, that each exert a repulsive or attractive force). Modeling dynamic obstacles (e.g., intruding aircraft, or Temporary Flight Restriction) or sensor noise requires the transformation of precise position or velocity to be represented as probability distributions and causes an ever-increasing computational complexity for the algorithm. At its root, RRT assumes deterministic state transitions, and research into adaptations of RRT has provided solutions to include uncertainty; however, computation time may still be an issue when re-planning in real time \cite{jiang2021r2}. Playbooks derived from MILP optimization have been extensively investigated, but can also suffer from the computational time required and symmetric solutions requiring heuristics to break ties~\cite{saber2023optimized,saber2023robust}

In recent years, Reinforcement Learning (RL), a subset of Machine Learning (ML), has shown equivalent or better performance than previously introduced path planning techniques by directly learning a policy through interaction with a simulated environment that contains uncertainty in its modeling. A key enabler leading to the rise in the use of RL to solve challenging problems is in part due to the ability to synthetically create large amounts of data through simulated environments. Specifically, in sequential decision-making problems, a RL approach incorporates uncertainty into the learning process by modeling the environment as a Markov Decision Process (MDP) and allowing agents to learn through interaction with the environment. In aviation, RL has seen applications in air traffic management, for maintaining the separation distance between aircraft (known as conflict resolution and separation assurance), scheduling, capacity balancing, and Detect and Avoid (DAA)~\cite{razzaghi2022survey}.

Inspired by these achievements, the authors had two primary objectives: (1) to create a simulation environment or framework in which AI-based CM techniques may be evaluated, and (2) to implement and evaluate an initial set of techniques from literature. A prior publication focused on creating the simulation environment and framework~\cite{alvarez2023towards}. In this paper, we focus primarily on the second objective, extending the prior simulation framework and evaluating trained AI-based versus heuristic path planning techniques. The remainder of the paper is organized as follows. Section \ref{sec:background} discusses background information on RL in aviation and their associated challenges. Section \ref{sec:ProblemFormulation} outlines the formulation of the training and evaluation simulation environment. Section \ref{sec:algorithm_design} provides a description of the algorithm design and reward modeling of the agents under test. Section \ref{sec:Results} discusses results from the evaluation of each contingency management agent under test. Section \ref{sec:Conclusion} provides a summary of observations made during training and evaluation of the agents. Section \ref{sec:FutureWork} makes recommendation on additional work necessary to refine the agents developed and path forward.

\section{Background}
\label{sec:background}
Prior research has demonstrated Deep Reinforcement Learning (DRL) systems can outperform classical approaches in a variety of complex decision-making tasks. Unlike heuristic-based systems, which use predefined rules and expert knowledge, DRL systems learn and adapt to new situations through experience and interaction with the environment. As an example, to support traffic separation assurance in commercial aviation, researchers have implemented a variety of approaches to train AI agents to resolve traffic conflicts. Each implementation models uncertainty as an MDP where the actions are composed of speed changes, lateral movements, or vertical flight level changes. The defining differences between implementations, outside of simulator and uncertainty modeling, are the structure of the state space, action space, and algorithm used for the learning-based logic. Moreover, when compared to MILP approaches, which are limited to offline execution due to extensive computational resources and are constrained by their formulation, DRL systems offer more flexible and scalable solutions. These advantages make DRL a promising approach for addressing the dynamic and unpredictable nature of real-world problems~\cite{D2MA, D2MAV, pham2019reinforcement, wang2019deep, hong2014application}.

In the case of DAA, the Federal Aviation Administration (FAA) introduced the use of machine learning approaches in safety critical systems within the next generation family of collision avoidance systems known as Airborne Collision Avoidance System X (ACAS X) for commercial (ACAS Xa), large unmanned (ACAS Xu), small unmanned (ACAS sXu), and rotor-craft (ACAS Xr) \cite{Kochenderfer2013, owen2019acas, alvarez2019acas, speed}. This family of collision avoidance systems modeled the collision problem as an MDP where the transition model between states is fully defined and Q-Tables are solved through value iteration. Due to the ``curse of dimensionality'' increasing the memory footprint of these Q-tables, the FAA TCAS Program Office also conducted research into table compression and fully defining the three-dimensional collision problem through neural network implementations. However, these solutions were unable to achieve the equivalent safety and alerting balance as the published standards with the same training environment and encounter modeling~\cite{julian2016policy,  corteguera2020airborne, CooperDRL}. Alternative research approaches have used Proximal Policy Optimization (PPO), Deep Q-Networks (DQN), and double deep Q-network (DDQN) which have also shown promising results in providing collision free trajectories with a return to waypoint optimization strategy~\cite{li2019optimizing, panoutsakopoulos2022towards}. 

Due to the promising results observed in prior studies, the authors set out to meet two key objectives:

\begin{enumerate}
    \item Create a simulation environment and overarching framework for rapidly developing, evaluating, and verifying AI/ML-based techniques as applied to CM for future In-time Aviation Safety Management Systems (IASMS)~\cite{ellis2021concept}.
    \item Use the framework/tool to evaluate a specific set of AI/ML techniques as part of specific CM designs and associated performance metrics.
\end{enumerate}

If achieved, this will help lead to research-informed standards for simulator(s)/testbed(s) and performance metrics to evaluate AI approaches that may be developed in academia, government, and industry. This may also help to standardize the training and validation of AI/ML-based algorithms for other functions implemented in aviation~\cite{Agogino2024NASA}.

Prior work has led to an initial development framework (i.e., the AAM-Gym testbed~\cite{aamGymMarc, alvarez2023towards}) that is extended here. AAM-Gym allows the user to rapidly prototype, simulate, and analyze use-cases where novel AI/ML-based safety management solutions are being investigated. In this work the use-case (or application) is automated risk mitigation (i.e., auto-CM) for representative AAM/UAM scenarios. AAM-Gym contains a set of AI/ML-based algorithms that may be used as-is or tailored by the user, and it allows for inclusion of new algorithms or logic primitives.
Once AL/ML-based algorithms are selected (or defined), the framework allows the user to choose a set of performance metrics of interest; define assumed human roles and procedures (if any); and define the degree of interaction with DAA or other automation that may independently mitigate other types of hazards not directly considered by the CM design.
The framework allows for consideration of AI/ML-techniques both as a means of performing CM, and as a tool to help evaluate other means of implementing CM (e.g., more traditional structured logic-based methods). For example, in this paper we use AAM-Gym to help assess various means of choosing and executing contingencies, and we use AAM-Gym to help assess various AI/ML techniques to help in understanding their limitations and benefits. Lastly, the framework and simulation tests can help in understanding interoperability issues within systems of systems (e.g., DAA and CM operating together); this topic is to be addressed in future work. We feel such a framework is critical to comprehensive verification and validation of AI/ML-based applications used in safety-critical aviation systems.

 Because of the prior work and lessons-learned on applications of AI/ML to DAA, it seemed a good starting point for the work described here on CM. Namely, testing a simulation capability and evaluating an initial set of DRL-based techniques. The remainder of this paper focuses on work toward Objective 2; however several lessons-learned during implementation and testing have led to improvements in the framework (Objective 1).

\section{Problem Formulation}
\label{sec:ProblemFormulation}
This section explains the approach to establish the algorithms, models, scenarios, and metrics used for the testing of the selected CM agents.

\subsection{DRL Algorithms}
Basing the DRL contingency management framework on prior AAM-Gym developments~\cite{aamGymMarc, alvarez2023towards} allowed the research team to quickly adapt and test algorithms that adopt the OpenAI~\cite{https://doi.org/10.48550/arxiv.1606.01540} interface. Two algorithms, D2MAV~\cite{D2MAV_A} and Soft Actor Critic with Discrete actions (SACD)~\cite{sacd} extended with attention, are tested for training a DRL agent. These algorithms have demonstrated efficacy in supporting separation assurance in AAM environments~\cite{D2MA, D2MAV, MAASA, brittain2023improving}. These algorithms were initially used in the development of the simulation framework and described in~\cite{alvarez2023towards}.

\subsection{Baseline Simulation and Heuristic Algorithm}
As baselines, this paper uses a heuristic based agent inspired by \cite{doi:10.2514/6.2022-3459} and nominal simulations where no agent interacts with the aircraft. The heuristic based agent has equivalent state space information as the DRL with the additional knowledge of the nodes that define the corridor network. The heuristic based agent uses estimates of the energy reserves during flight and constantly monitors the energy required to continue the flight plan as well as the probability of loss-of-control\footnote[1]{For the tested scenarios, we use the term loss-of-control to represent the case where there is degraded navigation system performance such that the vehicle is no longer able to navigate to the next waypoint within an acceptable safety margin.}. The criteria for taking control of the aircraft is defined by Algorithm \ref{alg:heuristic} in the Appendix and is summarized as follows:

\begin{enumerate}
    \item Calculate the remaining flight time available and required flight time to continue route, based on energy reserves.
    \item Reroute aircraft if the energy required to complete flight plan is greater than energy reserves.
    \item Reroute aircraft if it experiences or encounters any hazard along the flight path.
    \item Reroute aircraft if there is a hazardous situation at the destination that would preclude landing (i.e., $P_{H_{dest}}$ is greater than zero).
    \item Reroute utilizing corridor network if aircraft can sustain flight, and there is no hazard found along the path.
    \item If flight path through corridor network is not possible, reroute directly to closest vertiport whose direct path does not cross a hazardous area or has the lowest total risk.
\end{enumerate}

\subsection{Hazard Modeling}
The framework allows for the introduction of a variety of hazard types that may impact aircraft dynamics and state. In this paper the hazards of interest are 1) wind, 2) no fly zones, 3) regions of high-density population (modeled as risk fields) 4) increased energy consumption due to the charge cycle of the battery, and 5) navigation system performance where aircraft experience a loss-of-control. 3D wind fields are generated using a back-end simulator (BlueSky [38]). For training initial tasks, a constant zero-wind is used. Then, as tasks and training get more complex, randomly generated winds fields are used. This is further discussed in Sec \ref{sec:ExperimentSetup}. A New York City population density model is extracted from the Oak Ridge National Library population database and merged with the temporal component of the New York City Taxi and Limousine Trip Record Data \cite{OakRidgePopulation, NYCTaxiData}. This population data is used with a ballistic trajectory model, defined by~\cite{doi:10.2514/6.2022-3459}, to estimate the probability of casualties when aircraft reaches zero energy. 

\subsection{Aircraft Energy Modeling}
The aircraft available energy is modeled by the linear energy model defined in \cite{alvarez2023towards}. The model is modified to directly initialize the aircraft with an energy level defined by a uniform distribution [20, 250] KWh. The energy model uses three properties to calculate the energy consumed for a given time step: 1) the battery life cycle for the battery assigned to the aircraft, 2) a constant energy consumption rate, and 3) a battery failure probability assigned to each aircraft. The probability of battery failure is an additional feature added to the energy model to further introduce uncertainty in the likelihood of aircraft reaching their desired destination even with high energy levels assigned at the beginning of flight. During agent training these factors on average lead to total flight times between 2--41 minutes. This range in flight time allows the DRL algorithms to experience complete energy failure of the aircraft and learn to divert aircraft to alternative vertiports or return to the departure vertiport. During evaluation of the algorithms, the range of initial energy is modified to explore the behavior learned.

\subsection{Scenario Generation}
Training and evaluation of DRL agents requires an accrual of experience by interacting with a simulation environment to learn the optimal policy. For the testing reported in this paper, the environment is initialized by randomly sampling from 40 days of synthetic AAM traffic generated with UAMToolKit, an event driven AAM traffic capacity and demand modeling tool~\cite{UAMToolKit}. The toolkit assumes a total fleet size of 100 aircraft. Since the focus of this research is on the cruise/enroute phase of the flight, the UAMToolKit parameters for capacity at vertiports, passenger capacity, turn-around times, and wind data are selected to maximize the number of airborne aircraft. Thus, the constraint on vertiport and passenger capacity is removed by setting it to unlimited capacity. While wind field information is ignored for traffic generation, and a one-minute turn-around time is selected to maximize aircraft utilization. All operations are uniformly distributed in between eight altitude lanes from 1,000-5,000 ft AGL traversing between vertiports through a pre-existing helicopter network designed by~\cite{UAMToolKit}. A uniform distribution is used to assign an aircraft type with its associated dynamic limits from a set of representative UAM vehicles whose names are obscured for anonymity. Table \ref{tab:aircraft_dynamics}, summarizes the aircraft dynamics of each vehicle type.

\begin{table}[tbh]
\caption{Vehicle power and speed ranges.}
\centering
\label{tab:aircraft_dynamics}
\begin{tabular}{lll}
\hline
Aircraft Type & Battery Charge (KWh) & Speed Range (knots)  \\ \hline
UAM-A & 20 - 250  & 20 - 174 \\
UAM-B & 20 - 250  & 20 - 156 \\
UAM-C & 20 - 250  & 20 - 148 \\
UAM-D & 20 - 250  & 20 - 130  \\
\hline
\end{tabular}
\end{table}

\subsection{Metrics} 
All aircraft state and environment information was recorded for post processing and evaluation. In addition, the following metrics were calculated during run time: total reward experienced by each aircraft, aircraft reaching each terminal state (i.e., reaching destination vertiport, rerouted to alternate vertiport, return to departure vertiport, experience loss-of-control, energy depleted), time spent in flight route, contingency actions taken, latitudes and longitudes traversed, and the projected probability of casualty when reaching the terminal state. 

\section{Algorithm Design}
\label{sec:algorithm_design}
This section discusses design and training of the CM agents using two previously developed DRL algorithms. We extend a previous DRL framework~\cite{alvarez2023towards} by introducing additional state information, rewards, and an implementation of a heuristic based approach.

\subsection{Curriculum Training}
Undertaking the development of a DRL agent requires understanding the use case objectives and fine tuning each reward parameter to achieve the desired behavior. Inspired by the human learning process, we adopt a curriculum learning strategy where the agent is sequentially trained through a set of increasingly complex tasks supportive of CM for UAM operations. Each task builds upon the previous one, allowing the agents to learn robust contingency management strategies, as shown in Table~\ref{tab:curriculum}.

Each agent is pre-loaded with a set of possible vertiport landing sites and the planned destination. At each time step in a simulated flight, the agent must decide whether to intervene and reroute the aircraft to an alternate vertiport, return to the departure vertiport, or continue to its original destination. In this paper, the factors that influence an agent's CM decisions are available power, wind, navigation system performance, no-fly zones, and underlying population density. 

\begin{table}[h]
    \centering
    \caption{Curriculum Learning Tasks.}
    \label{tab:curriculum}
    \begin{tabular}{c | l}
        \hline
         Task & Description  \\ \hline
         T1    & Agent learns to reach destination. \\ 
         T2    & T1 and selects to reroute to alternative vertiport due to reduced energy levels. \\
         T3    & T2 and is cognizant of probability of loss of aircraft control fields and no-fly zones. \\
         T4    & T3 and wind fields are included in the simulations with agent cognizant of wind speeds at current location. \\
         T5    & \multirow{2}{15cm}{T4 and is cognizant of population density and selects to navigate through a low population zone as flight conditions worsen.} \\
    \end{tabular}
\end{table}

\subsection{State Space}
Each aircraft, referred to as ownship, in the simulation environment is controlled by an independent instance of the contingency management agent. The ownship's state $s$ at time $t$ is defined by Eq~\ref{eq:state_space}.

\begin{equation}
\label{eq:state_space}
    s_{t} = O_{t} + V_{t} + h_{t}
\end{equation}

\begin{equation}
    O_{t} = (\psi, z, \dot{v_{xy}}, v_{xy}, E, \dot{E}, P_c, W_y, W_x, (x^{(j)}_{\text{wpt}}-x, y^{(j)}_{\text{wpt}}-y, P^{(j)}_H, \rho^{(j)}_{dest}) \quad \forall \; j \in [1, N_{\text{wpt}}]
\end{equation}

where $O_t$  represents ownship's heading ($\psi$), altitude ($z$), horizontal acceleration ($\dot{v_{xy}}$), horizontal true air speed ($v_{xy}$), remaining energy ($E$), and energy used within a time step ($\dot{E}$). $P_c$ is the probability of casualties in the event of loss of aircraft control due to complete battery failure or a loss-of-control field. $W_y$ and $W_x$ represent the north and east wind speed respectively. Additionally, $N_{\text{wpt}}$ number of waypoints along the original flight plan are provided with relative east (X) and north (Y) location to ownship $(x^{(j)}_{\text{wpt}}-x, y^{(j)}_{\text{wpt}}-y)$, hazard function output at waypoint ($P^{(j)}_H$), and distance to destination from waypoint ($\rho^{(j)}_{dest}$).

\begin{equation}
    V_{t} =  \theta_{dest}, \rho_{dest}, P_{H_{dest}}, (P^{(j)}_{H_\theta}) \quad \forall \; j \in [1, N_{\theta}+1],
\end{equation}

$V_{t}$ contains the ownship relative bearing to destination ($\theta_{dest}$), distance to destination ($\rho_{dest}$), hazard function output at destination ($P_{H_{dest}}$), and $P^{(j)}_{H_\theta}$, the hazard function output of $N_{\theta}$  points within a radius \( R_{field} \) from the ownship, such that these points are evenly distributed in a circle around ownship and an additional point at the current location of ownship. The angle between each of these adjacent points is given by:
\[
\Delta\theta = \frac{360^\circ}{N_{\theta}}
\]

\begin{equation}
     h_{t} = (\rho^{(i)}, \theta^{(i)}, P^{(i)}_H) \quad \forall \; i \in [1, N_{vertiports}]
\end{equation}

The vertiport state information $h^{(i)}$ at time $t$ includes the relative distance from ownship to the vertiport ($\rho^{(i)}$), relative bearing to ownship ($\theta^{(i)}$) and probability of loss of aircraft control at the vertiport ($P^{(i)}_H$), defined by a hazard function.

\subsection{Reward Model}
In prior research we presented a reward model formulated for contingency management DRL agents \cite{alvarez2023towards}. This section discusses extensions of the reward model and the observed or expected behavior that led to changes in the reward model. Due to the curriculum training approach undertaken, not all rewards are active in each curriculum task, and their specific use will be highlighted in the results. The action space is extended from prior research to include two additional heading change actions of $\pm$1$\degree$ for the agent to fine tune the desired heading towards a vertiport, see Table \ref{tab:action_sets}. \\

\begin{table}[h]
\caption{Actions available for agents.}
\centering
\label{tab:action_sets}
\begin{tabular}{ll}
\hline
Description & Value\\ \hline
Heading change &  -5\degree, -1\degree, 0\degree, 1\degree, 5\degree       \\
No action &  True, False     \\
Use assigned flight route & True, False   \\
\hline
\end{tabular}
\end{table}

The available actions are defined as follows:

\begin{enumerate}
    \item Heading change - The agent commands a left or right turn by a pre-defined magnitude (i.e., $\pm$5$\degree$, $\pm$1$\degree$, 0$\degree$) and continues this heading change until a new action is chosen.
    \item No action - The agent does not command an action and the aircraft continues on its current heading or active flight plan.
    \item Use assigned flight route - The aircraft continues on its planned flight route or returns to the nearest waypoint of the flight plan, if prior actions caused a deviation, and continuous along flight route.
\end{enumerate}
\vspace{2mm}
The reward function utilized by the MDP is defined by Eq.\ref{reward_func}, where $R(s_{t})$, $R(a_{t})$ are defined by Eq.~\ref{eq:r_st},~\ref{eq:r_at}.  
\begin{equation}
\label{reward_func}
R(s_t, a_t, p_t, t) = R(s_t) + R(a_{t}),
\end{equation}

\begin{equation}
\label{eq:r_st}
R(s_t) = 
      \delta{T_{energy}}+\delta{T_{H}}+\delta{T_{vertiport}}+\delta{T_{population}+\Omega}
\end{equation}

$R(s_t)$ encompasses the state-dependent penalties and rewards triggered by an aircraft reaching a specific state, as defined by Eq.~\ref{eq:r_st}. One penalty is applied when an aircraft's energy reserves, $E$, are below 90 KWh or completely depleted, as defined by Eq.~\ref{eq:energy_penalty}. In addition, when an aircraft's energy reserves are depleted, the aircraft is terminated, and also incur a penalty of $\delta{T_{population}}$, defined by Eq.~\ref{eq:r_population}. When an aircraft enters a loss-of-control field (i.e. navigation performance hazard) or no-fly zone the probability of terminating the flight is calculated with a hazard function and compared against a zero-mean Gaussian distribution or a threshold criteria and incurs a penalty of $\delta{T_{H}}$, as defined by Eq.~\ref{eq:r_loss_control}. Where $D_H$ is the distance of the aircraft to the center of the hazard.

\begin{equation}
\label{eq:energy_penalty}
    \delta{T_{energy}} =
    \begin{cases}
        -2.0 & \text{if $E = 0.0$ KWh} \\
        -0.0015 & \text{if $E < 90.0$ and $E > 0.0 $ KWh} \\
        0 & \text{else} \\
    \end{cases}
\end{equation}

\begin{equation}
\label{eq:r_loss_control}
    \delta{T_{H}} = \\
    \begin{cases}
        -0.24 & \text{if $D_H$ < 1,000 m and $P_{H_{center}} > \beta_{H}$} \\
        -0.12 & \text{if $D_H$ < 1,000 m and $P_{H_{center}}\leq \beta_{H}$}\\
        0   & \text{else} \\
    \end{cases}
\end{equation}

If the hazard region is representing a loss-of-control field, the value of Eq.~\ref{eq:p_h_function} is compared against Eq.~\ref{eq:zmg} (i.e., Eq.~\ref{eq:p_h_function} > Eq.~\ref{eq:zmg}) to assess whether the flight should be considered as terminated. In addition, the value of $\beta_H$ in Eq.~\ref{eq:r_loss_control} takes on the output of Eq.~\ref{eq:zmg} and $P_{H_{center}}$ is the value of Eq.~\ref{eq:p_h_function} at the aircraft location. The process is equivalent if the hazard region is representing a no-fly zone, with $\beta_H$ taking on the value of $\beta_{H_{no-fly}}$ to assess whether the flight should be considered terminated, and the penalty. Lastly, when an aircraft is within a specified range of a vertiport, $d_{goal}$, the aircraft is terminated and awarded $\delta{T_{vertiport}}$. The value $\delta{T_{vertiport}}$ is dependent on the vertiport reached: the destination, an alternative vertiport, or departure vertiport. The value of $\delta{T_{vertiport}}$ is calculated by Eq.~\ref{eq:d_vertiport} and Eq.~\ref{eq:delta_return_home}. In Eq.~\ref{eq:delta_return_home}, $T_{straight\_path\_dest}$ represents the time required to fly directly to the destination, and $P_{H_{straight\_path\_dest}}$ denotes the accumulated risk of flying directly to the destination.

The multivariate Gaussian distribution defining the hazard region is dependent on the relative position of the aircraft to the center of the distribution, as seen in Eq.~\ref{eq:p_h_function} where the variables are in units of meters defined in the local east, north, up coordinate system.

The population penalty incurred during energy depletion is calculated by 1) projecting the aircraft impact location assuming a ballistic trajectory, 2) querying the population database for the expected individuals at projected location and 3) assuming a uniform distribution of population within the 100x100 m bin provided by the population database to output a probability of casualties $P_c$, and the fraction of the population impacted $P_p$. The penalty is then calculated using Eq.~\ref{eq:r_population}. Lastly, a step penalty, $\Omega$, is applied that decreases as the aircraft approaches the destination, defined by Eq.~\ref{eq:step_penalty}.

\begin{equation}
\label{eq:p_h_function}
P_H(X,Y) = e^{\frac{(X-\mu_{xi})^2+(Y-\mu_{yi})^2}{-2\sigma^2}}
\end{equation}

\begin{equation}
\label{eq:zmg}
    \beta_{H_{loss}} = \mathcal{N}(\mu=0.0, \sigma=0.26903^2)
\end{equation}

\begin{equation}
\label{eq:d_vertiport}
\delta_{vertiport} = \\
    \begin{cases}
        1.0 & \text{if vertiport = destination}\\ 
        0.5 & \text{if vertiport $\ne$ destination and $\ne$ takeoff location}\\ %
        \delta_{vertiport\_return\_home} & \text{if vertiport = takeoff location}\\ %
    \end{cases}.
\end{equation}

\begin{equation}
\label{eq:delta_return_home}
\delta_{vertiport\_return\_home} = \\
    \begin{cases}
        -\frac{E}{1000} & \text{if $ \frac{E}{\dot{E}} > T_{straight\_path\_dest}$ and $P_{H_{straight\_path\_dest}}==0 $} \\ 
        0.5 & \text{else}\\ %
    \end{cases}.
\end{equation}

\begin{equation}
\label{eq:r_population}
\delta{T_{population}} = \frac{-1}{\log\left(\frac{b - 1}{b}\right)} \log(\frac{b - P_cP_p}{b})
\end{equation}

\begin{equation}
\label{eq:step_penalty}
    \Omega =
    \begin{cases}
        -\delta_{step}\cdot e^{\frac{-4 \cdot \rho_{dest}}{d_{max}}} & \text{if $\rho_{dest} \leq d_{max}$} \\
        -\delta_{step}\cdot e^{-4} & \text{else} \\
    \end{cases}
\end{equation}

$R(a_{t})$ is defined by
\begin{equation}
\label{eq:r_at}
R(a_{t}) = \\
    \begin{cases}
      \delta_{action\_penalty} & \text{if $a_{t} \neq$ ``No action'' or $a_{t}\neq$''Use assigned flight route''} \\
      0 & \text{else} \\
    \end{cases},
\end{equation}
where $\delta_{action\_penalty}$ is a penalty applied when any action is chosen other than ``No action'' or ``Use assigned flight route''.

 The final reward model parameters, as shown in Table~\ref{tab:reward_parameters}, were individually tuned by using a manual coordinate descent search.

\begin{table}[H]
\caption{Final Reward Parameters and Values.}
\centering
\label{tab:reward_parameters}
\begin{tabular}{lll}
\hline
Parameter & Value \\ \hline
$N_{\text{wpt}}$ & 2 \\
$N_\theta$ & 20\\
$N_{vertiports}$ & 29  \\
$d_{\text{max}}$ & 647,391.47 m \\
$\delta_{\text{action\_penalty}}$ & -0.001 \\
$\delta_{\text{step}}$ & -0.0001 \\
$\delta{T_{energy}}$ & -2.0 \\
$\delta{p_{max}}$ & -0.00015 \\
$\sigma$ & 269.023 \\
$\beta_{H_{no-fly}}$ & 0.2 \\
$P_{H_{threshold}}$ & 0.2 \\
$b$ & 0.001 \\
$R_{field}$ & 500 m  \\
$d_{goal}$ & 500 m \\

\hline
\end{tabular}
\end{table}

\subsection{Experiment Setup}
\label{sec:ExperimentSetup}
A span of tests were designed to help understand and advance the overarching framework (i.e., Objective 1) as well as to evaluate the set of CM agents previously described (Objective 2). The Lincoln Laboratory Supercomputer Center (LLSC) was utilized in this study to meet the substantial data requirements for training and evaluating the DRL agents. The compute architecture comprised 16 Intel Xeon Gold 6248 2.5 GHz compute nodes, each equipped with two NVIDIA Tesla V100 graphics processing units (GPUs)~\cite{reuther2018interactive}. Each curriculum task listed in Table \ref{tab:curriculum} necessitated the careful introduction of rewards and state inputs to achieve satisfactory performance. This performance was attained by training a DRL agent with 40 distinct parallel simulations, each independently running 3-hour windows of operations. Training at each curriculum task involved 20,000 iterations, with each iteration consisting of 64 training steps (equivalent to 5 simulation seconds) per 40 distinct parallel simulation. Consequently, each curriculum task required an average of 68 hours of wall-clock time.

\begin{figure}[tbh]
\begin{center}
\centerline{\includegraphics[width=0.85\textwidth]{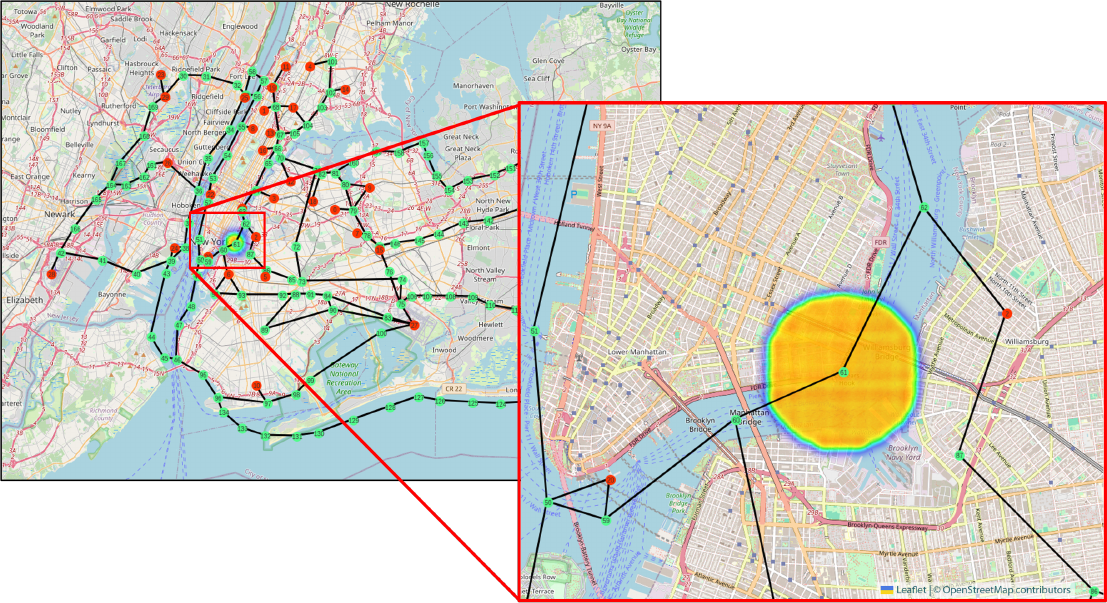}}
\caption{Network utilized for evaluation with loss-of-control hazard to force the majority of nominal traffic to traverse through hazard. Vertiport locations displayed with red circles, and nodes defining network segments displayed with green circles. Colormap indicates the distribution of hazard severity within the hazard region (red, more severe; blue, less severe).\textsuperscript{*}}
\label{fig:hazard_location}
\end{center}
\end{figure}

In curriculum task T3, the centers of the loss-of-control and no-fly zone hazard regions are randomly initialized to align with a node in the network as designed in \cite{UAMToolKit}. Wind speed and direction are randomly sampled from independent uniform distributions ranging in [0, 10] knots and [0, 360] degrees. The velocity of aircraft is initialized using a uniform distribution with velocities ranging between [5, 65] knots. A total of 100 simulated days of operations are used for evaluation of each algorithm. Each evaluation contains a maximum of 100 aircraft simultaneously operating in the airspace. The results show the mean value of the metrics and two standard deviations ($\sigma_2$) from the mean (bootstrapped). The loss-of-control hazard region is held static in the evaluation to force a majority of the nominal traffic to traverse through the hazard, as seen in Figure \ref{fig:hazard_location}.

\section{Results}
\label{sec:Results}
\begin{figure}[H]
    \centering
    \includegraphics[width=\textwidth]{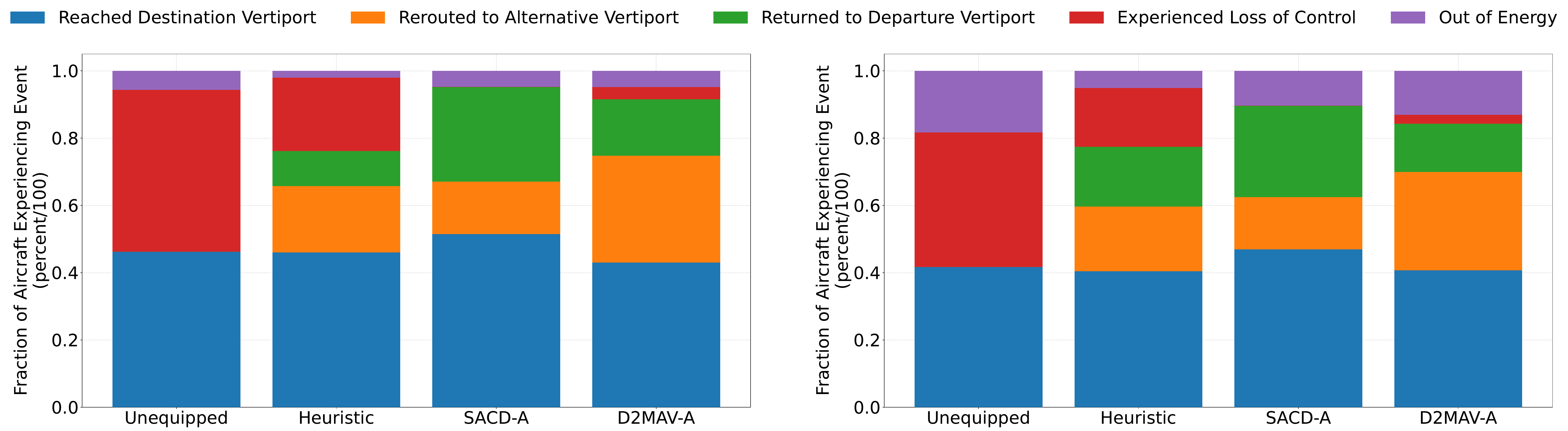} 
    \begin{minipage}[t]{0.45\textwidth}
        \subcaption{Evaluations with full range of energy reserves.} 
    \end{minipage}
    \hfill
    \begin{minipage}[t]{0.45\textwidth}
        \subcaption{Stressing evaluations with reduced energy reserves.}
    \end{minipage}
    \caption{Comparison of the average performance of 100 evaluations with the heuristic, SACD-A, D2MAV-A, and unequipped agents. Subplot (a) includes aircraft initialized with 20-250 KWh, while (b) includes only aircraft initialized with 20-90 KWh.}
    \label{fig:combined}
\end{figure}

\begin{figure}[H]
    \centering
    \includegraphics[width=\textwidth]{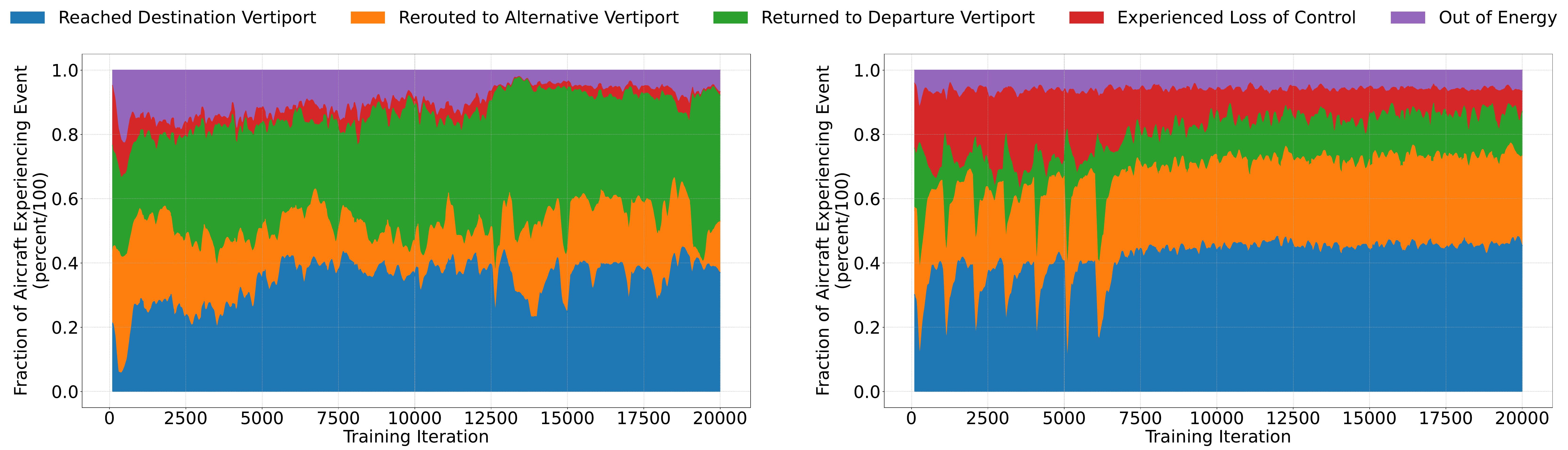}
    \begin{minipage}[t]{0.45\textwidth}
        \subcaption{Curriculum T4 learning curves of SACD-A agent.}
    \end{minipage}
    \hfill
    \begin{minipage}[t]{0.45\textwidth}
        \subcaption{Curriculum T4 learning curves of D2MAV-A agent.}
    \end{minipage}
    \caption{Final training of (a) SACD-A, and (b) D2MAV-A agents for curriculum task T4 where wind is introduced. High fraction of aircraft reaching a vertiport ($\sim$0.8) at iteration 0 is due to the application of transfer learning from previous curriculum tasks. Rolling mean of 100 iterations used to remove noise in plotting of data.}
    \label{fig:training_example}
\end{figure}

\begin{figure}[H]
    \centering
    \includegraphics[width=\textwidth]{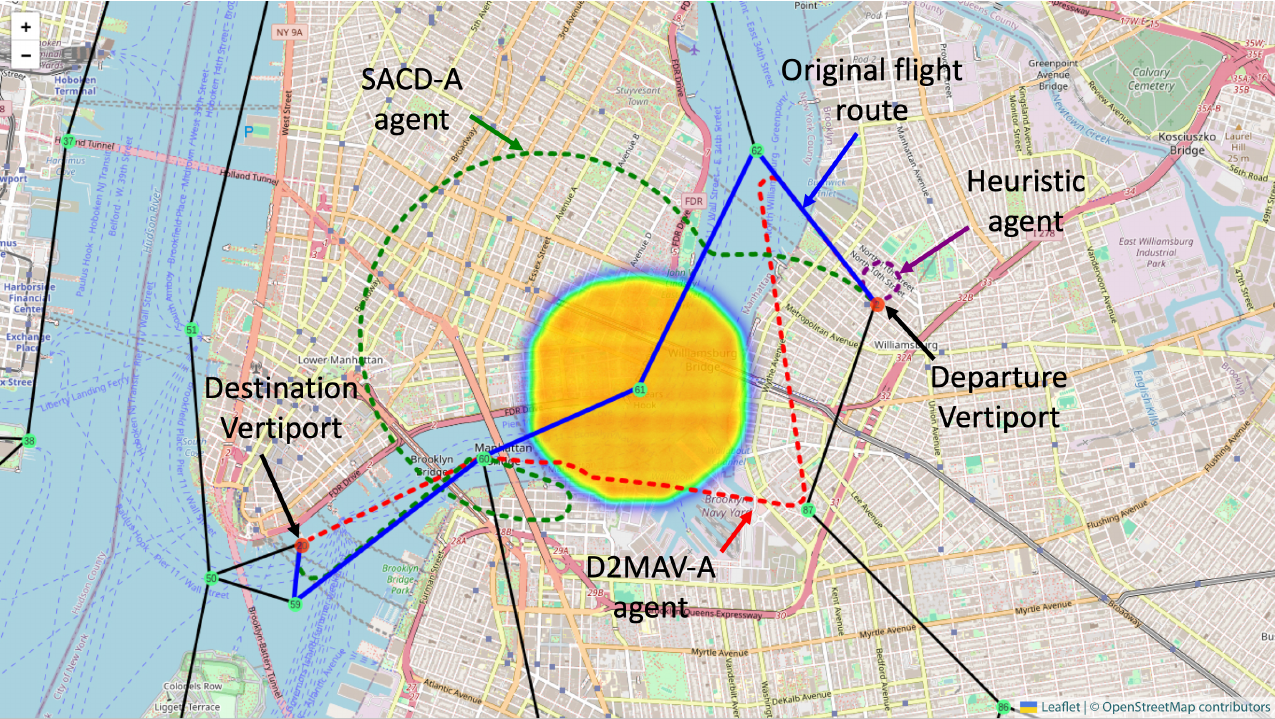}
    \caption{Example case where DRL reroutes may be operationally unsuitable.\textsuperscript{*}}
    \label{fig:track_examples}
\end{figure}

\begin{figure}[H]
\begin{center}
\centerline{\includegraphics[width=0.8\textwidth]{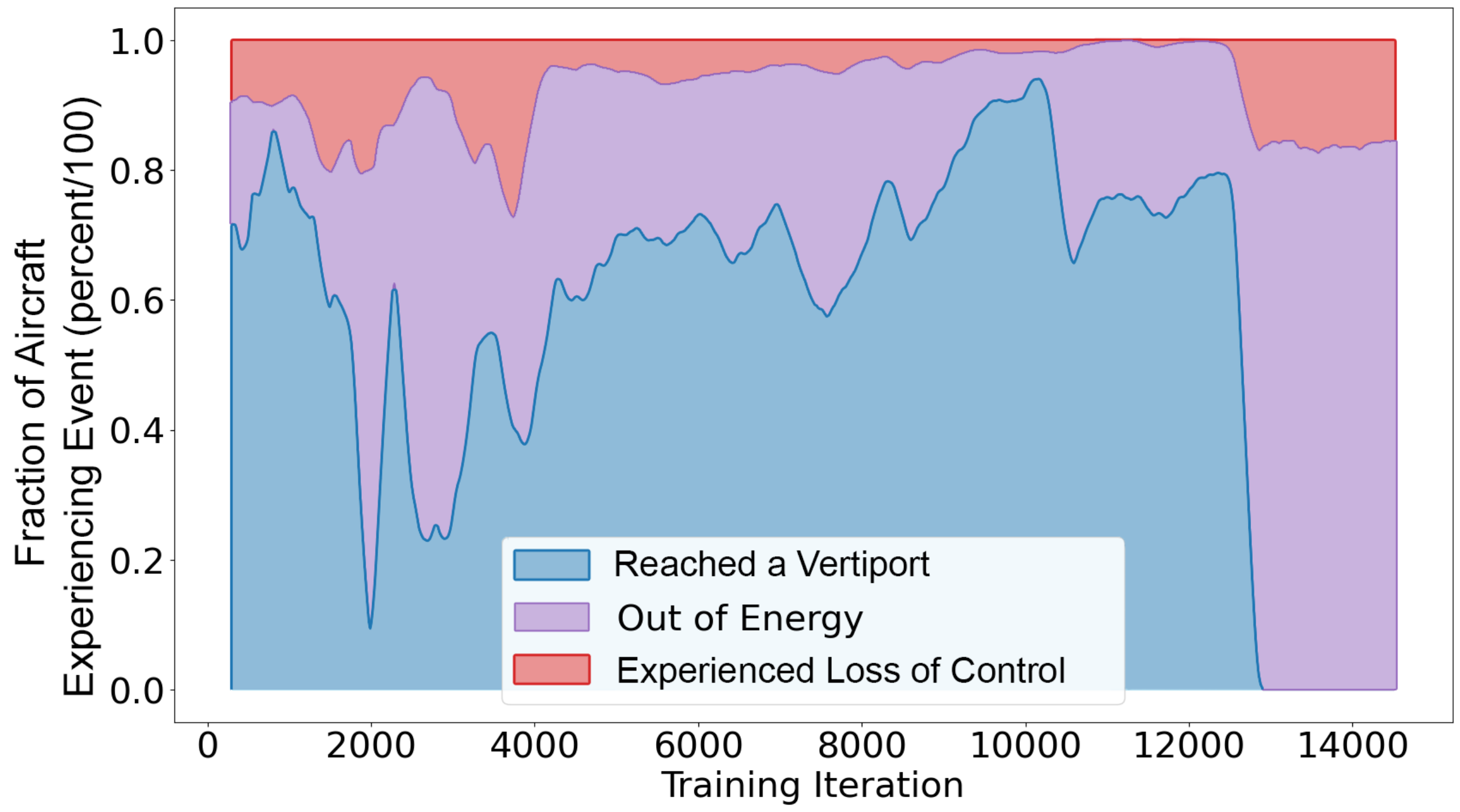}}
\caption{Example of catastrophic forgetting experienced during transition from T2 to T3 curriculum task training at iteration $\sim$13000.}
\label{fig:catastrophic_forgetting}
\end{center}
\end{figure}

\footnotetext{Map data from OpenStreetMap openstreetmap.org/copyright}
In this section we discuss the observed performance of each agent versus nominal operations (i.e. unequipped). Figures \ref{fig:combined} to \ref{fig:catastrophic_forgetting} summarize collected data; which includes, for each agent, 100 evaluations repeatedly simulated with a unique random seed used for all agents. Over 798 million operations providing over 266 million hours of flight data were collected for the DRL agents to achieve the performance through the curriculum learning. Overall when the aircraft fleet is initialized with energy reserves sampled from a uniform distribution equivalent to the training scenarios, 20--250 KWh, the DRL based agent successfully reroutes an average of 95.2\% ($\sigma_2=~\pm$~1.08\%) and 91.6\% ($\sigma_2=~\pm$~1.29\%) of aircraft to vertiports, for SACD-A and D2MAV-A respectively. The heuristic agent outperforms the nominal (unequipped) operations, however it only successfully reroutes 76.1\% ($\sigma_2=~\pm$~6.07\%) to a vertiport as compared to (46.2\%) for the nominal operations. Closer inspection of Fig.~\ref{fig:combined}a reveals that each agent maintains a comparable number of aircraft reaching their destination as in the nominal case with SACD-A rerouting more aircraft to their destination if the original flight plan is unfeasible.  Since all agents only have $N_{\text{wpt}}$ of flight path information they must utilize state information near ownship to sense regions of loss-of-control (i.e., $P^{(j)}_H$, $P_{H_{dest}}$ as defined in Section \ref{sec:algorithm_design}). While the heuristic agent allows for tracks to use routes with a risk of loss-of-control below $P_{H_{threshold}}$ to reduce situations where no route is possible, more aircraft are found to experience loss-of-control than $P_{H_{threshold}}$ and reducing this value leads to a tradeoff in more aircraft experiencing energy depletion due to the longer flight paths selected. This tradeoff is due to a lack of global information resulting in sub-optimal routing causing 21.8\% ($\sigma_2=~\pm$~5.612\%) of aircraft to experience loss-of-control.  The SACD-A nearly eliminates loss-of-control risk with only 0.07\% ($\sigma_2=~\pm$~0.119\%) of aircraft encountering it. This is achieved by SACD-A exploiting the state space information in training, Fig.~\ref{fig:training_example}, and rerouting aircraft around the hazard region, as seen in Fig. \ref{fig:track_examples}. Even with intensive training, Fig. \ref{fig:track_examples} shows the flight trajectories produced by the DRL agents can contain unexpected characteristics that may not make them operationally suitable. The sampling efficiency of SACD-A also leads it to outperform the D2MAV-A algorithm, a similar outcome to prior comparison~\cite{brittain2023improving}. To further evaluate the differences in performance, an evaluation of scenarios where aircraft are initialized with 20--90 KWh was conducted (i.e., less energy reserves). Figure \ref{fig:combined}b demonstrates that the performance of the DRL agent is robust to the reduction in aircraft energy as the results are similar to those of Fig. \ref{fig:combined}a with an increase in aircraft experiencing an "Out of Energy" terminal state.

Although the DRL agents outperform the heuristic agent in terms of aircraft reaching the original destination and leading more aircraft to a vertiport, the reward function required careful tuning to balance each additional objective as the agents progressed through the curriculum learning tasks. A known issue with DRL agents is called ``catastrophic forgetting'', where an agent loses previously acquired knowledge while learning a new objective. This phenomenon is prevalent in complex, dynamic environments, where continuous adaptation is required. This phenomenon was experienced between curriculum task T2--T3 when introducing loss-of-control regions as the logic transitioned from optimizing reduction in loss-of-control to further reduce the fraction of ``out of energy'' aircraft, as seen in Fig. \ref{fig:catastrophic_forgetting}. A common way to reduce the likelihood of catastrophic forgetting is to establish an experience replay, where a simulation buffers past experience~\cite{mnih2015human}. To reduce the instance of catastrophic forgetting, a replay buffer was implemented in this study, storing originally 4 million simulations steps when the phenomenon occurred, and increased to 8 million simulation steps thereafter. This was only possible due to the 364 GB of random-access memory available in the LLSC compute nodes. Moreover, during training of an agent, a rolling average of the reward function was used to assess the performance of the agent in training as well as a full evaluation of 10 days of simulation every 1000 training iterations. Network weights were only updated when these evaluation and training iterations would produce a higher average reward from a previous checkpoint.
\footnotetext{Map data from OpenStreetMap openstreetmap.org/copyright}

\section{Conclusion}
\label{sec:Conclusion}
A set of learning-based in-time decision-making agents were trained and evaluated as supportive of automation CM for AAM operations in the presence of hazards. The learning-based agents show DRL can meet complex safety objectives and are robust to uncertainty in the environment. When compared to a heuristic based approach, the DRL agents have a lower standard deviation in decision-making variability; with results within two standard deviations varying by less than ~2\% as compared to ~6\% indicating a higher level of understanding of the uncertainties in the environment. These results show promise for agents that can achieve both safety and operational efficiency objectives. To achieve the performance shown, over 798  million operations providing over 266 million hours of flight experience were collected for the DRL agents throughout the entire curriculum. This large amount of training experience highlights a tradeoff with expert knowledge-based heuristic systems. Careful tuning is also required to avoid catastrophic forgetting during training. Additionally, a notable tradeoff of using learning-based approaches is the challenge of achieving safety-critical assurance levels for ML-based agents. Rigorous verification and validation processes are needed, both for the agents themselves and the tools used to test them (e.g., the simulation environment). To help with this challenge, this work presents a framework to design, test, and evaluate a variety of algorithm approaches at scales previously not possible; and collect evidence such as suggested by~\cite{Agogino2024NASA}. As aviation moves towards increasingly autonomous systems, the development of comprehensive validation frameworks is essential to ensure safety and reliability.

\section{Future Work}
\label{sec:FutureWork}
Future research will focus on enhancing the robustness and efficacy of learning-based approaches in contingency management systems for AAM. Firstly, we will integrate  probability of casualty and the real-time risk assessment method suggested by~\cite{doi:10.2514/6.2022-3459} to the heuristic agent and compare the performance versus the DRL agents. Additionally, we intend to extend the range of hazards to include dynamic hazards and multiple hazards within the environment. Lastly, we aim to evaluate the interoperability of the agents with collision avoidance agents. This is needed to ensure seamless and appropriate coordination when independent autonomous agents may affect flight.

\section*{Acknowledgment}
The work was funded by NASA’s System-Wide Safety Project via an Interagency Agreement with the US Air Force (IA 80LARC23TA002, Project 10384).

\section*{Appendix}
\label{sec:appendix}

\begin{algorithm}[H]
\caption{Heuristic Agent Algorithm}
\label{alg:heuristic}
\begin{algorithmic}[1]
\State \textbf{Input:} Environment State Space
\State \textbf{Input:} Corridor Network Nodes (latitude, longitude, altitudes)
\State \textbf{Output:} Reroute or Continue Flight Path

\State Environment State Space: $S_t$ \Comment{see Eq. \ref{eq:state_space} for definitions}
\State Flight Time Remaining: $T_{remaining} \gets E/\dot{E}$ 
\State Flight Time Required: $T_{required} \gets \rho_{dest} / v_{xy}$
\State Probability of Loss of Control: $P_H \gets any(P^{(j)}_H >  P_{H_{threshold}}, P_{H_{dest}} >  P_{H_{threshold}})$
\State $IsReRouteRequired(S_t, network, T_{remaining}, T_{required}, P_H)$
\Function{RerouteInNetwork}{$S_t$, network}
    \State $\rho_{\text{min}} \gets \infty$  \Comment{Initialize the shortest route's distance to infinity}
    \State $r_{P_H} \gets \infty$  \Comment{Initialize the shortest route's probability of loss-of-control}
    \State $r_{selected} \gets None$  \Comment{Initialize route selected as None}
    \For{$vpt \in Veriports$}
        \State $R , R_{\rho}, R_{P_H} \gets Dijkstra(\text{closest node to aircraft location}, vpt, network)$ \Comment{Find route to vertiport $R$, path distance $R_{\rho}$ and output probability of loss-of-control of aircraft $R_{P_H}$ }
        \If{$R_{P_H} \leq P_{H_{threshold}}$ \textbf{and} $R_{\rho}<\rho_{\text{min}}$ } \
            \State $\rho_{\text{min}} \gets R_{\rho}$
            \State $r_{P_H} \gets R_{P_H}$
            \State $r_{selected} \gets R$
        \EndIf
    \EndFor
    \textbf{return} $r_{selected}$
\EndFunction
\Function{StraightReroute}{$S_t$, network} \Comment{Find shortest closest vertiport with no hazard in straight path}
    \State $\rho_{\text{min}} \gets \infty$  \Comment{Initialize the shortest distance to infinity}
    \State $r_{P_H} \gets \infty$  \Comment{Initialize path's probability of loss-of-control}
    \State $vpt_{selected} \gets None$  \Comment{Initialize route selected as None}
    \For{$vpt \in Veriports$}
        \State $R, R_{\rho}, R_{P_H} \gets Distance(\text{aircraft location}, vpt)$ \Comment{Straight route $R$, distance $R_{\rho}$ and Hazard in route $R_{P_H}$}
        \If{$R_{P_H} \leq P_{H_{threshold}}$ \textbf{and} $R_{\rho}<\rho_{\text{min}}$ } \
            \State $\rho_{\text{min}} \gets R_{\rho}$
            \State $r_{P_H} \gets R_{P_H}$
            \State $r_{selected} \gets R$
        \EndIf
    \EndFor
    \textbf{return} $r_{selected}$
\EndFunction
\Function{IsRerouteRequired}{$S_t$, network, $T_{remaining}$, $T_{required}$, $P_H$}
    \If{$T_{required}>T_{remaining}$ \textbf{or}  $P_H$}
        \State $r_{selected} \gets RerouteInNetwork(S_t, network)$
        \If{$r_{selected} \textbf{is} None$}
            \State $r_{selected} \gets StraightReroute(S_t, network)$
        \EndIf
        \State Reroute using $r_{selected}$
    \Else
        \State Don't Take Control
    \EndIf
\EndFunction

\end{algorithmic}
\end{algorithm}

\bibliography{sample}

\end{document}